\documentclass[authoryear]{elsarticle}
\usepackage{amssymb}
\usepackage{hyperref}
\usepackage{color,soul}
\usepackage{float}
\usepackage{enumitem}
\usepackage{afterpage}
\usepackage{color,soul}
\usepackage{xcolor}

\journal{Transactions in GIS}
\begin{document}
\hypersetup{hidelinks}
\begin{frontmatter}

\title{Extracting human emotions at different places based on facial expressions and spatial clustering analysis}

\author[1,2]{Yuhao Kang}
\author[2]{Qingyuan Jia}
\author[1]{Song Gao}
\author[2]{Xiaohuan Zeng}
\author[2,3]{Yueyao Wang}
\author[2]{Stephan Angsuesser}
\author[4]{Yu Liu}
\author[5]{Xinyue Ye}
\author[2]{Teng Fei}

\address[1]{Geospatial Data Science Lab, Department of Geography, University of Wisconsin, Madison, United States}
\address[2]{School of Resources and Environmental Sciences, Wuhan University, Wuhan, China}
\address[3]{College of Urban and Environment Sciences, Peking University, Beijing, China}
\address[4]{Institute of Remote Sensing and Geographical Information Systems, Peking University, Beijing, China}
\address[5]{Urban Informatics and Spatial Computing Lab, Department of Informatics, New Jersey Institute of Technology, Newark, United States}

\begin{abstract}
The emergence of big data enables us to evaluate the various human emotions at places from a statistic perspective by applying affective computing.
In this study, a novel framework for extracting human emotions from large-scale georeferenced photos at different places is proposed. After the construction of places based on spatial clustering of user generated footprints collected in social media websites, online cognitive services are  utilized to extract human emotions from facial expressions using state-of-the-art computer vision techniques. And two happiness metrics are defined for measuring the human emotions at different places. To validate the feasibility of the framework, we take 80 tourist attractions around the world as an example and a happiness ranking list of places is generated based on human emotions calculated over 2 million faces detected out from over 6 million photos. Different kinds of geographical contexts are taken into consideration to find out the relationship between human emotions and environmental factors. Results show that much of the emotional variation at different places can be explained by a few factors such as openness. The research may offer insights on integrating human emotions to enrich the understanding of sense of place in geography and in place-based GIS.
\end{abstract}

\begin{keyword}
affective computing \sep human emotion \sep place \sep big data  \sep GeoAI
\end{keyword}

\end{frontmatter}


\section{Introduction}
\label{S:intro}
Place, which plays a central role in daily life not only as a location reference  but also reflecting the way human perceive, experience and understand the environment, is a key issue in geography and GIScience \citep{tuan1977space,goodchild2011formalizing,winter2012approaching,scheider2014place,goodchild2015space,mckenzie2015poi,gao2017constructing,gao2017data,blaschke2018place,zhang2018representing,PurvesPlaces,wu2019fuzzy}. \citet{agnew2011space} proposed three aspects of place: location, locale, and the sense of place, which refers to the experiences of people and their perceptions and conceptualizations of a place. And place has been comprehensively depicted as the context and affordance with various human activities, which linked to memories and emotions of individuals \citep{jordan1998affordance,kabachnik2012nomads,scheider2014place,merschdorf2018revisiting}.
Human emotions, which are innately stored in human neural systems \citep{wierzbicka1986human,izard2013human}, provide bridges linking  the surrounding environments and human perceptions.
On one hand, emotions tint human experiences \citep{tuan1977space}, and show how places are psychologically felt by people \citep{davidson2004embodying}.
One the other hand, emotions are proved to be connected and affiliated with the surrounding things including living organisms \citep{wilson1984sociobiology}, nature environment \citep{capaldi2014relationship}, and cultural environment \citep{mesquita2004culture}, and so on. Therefore, understanding human emotions to the environment is important for human behavior analysis towards the sense of place \citep{grossman1977man,rentfrow2016geographical,smith2016emotion}.

Amount of early studies usually use questionnaires to investigate the emotion of people in different environmental contexts, which costs a lot of human resources and lacks timeliness \citep{golder2011diurnal}.
The emergence of big data and the development of information, communication, technology (ICT),  and artificial intelligence (AI) provide advanced methodologies and opportunities to solve the problems aforementioned in social sensing \citep{liu2015social,ye2016integrating,janowicz2019semantic}. 
Affective computing, as an interdisciplinary domain spanning computer science, psychology, and cognitive science, was proposed by \citet{picard1995affective} with a focus on investigating the interactions between computer sensors and human emotions. 
Every day, large volumes of geo-tagged user generated content (UGC) are uploaded to social networking websites such as Facebook and Twitter, photo-sharing sites such as Flickr and Instagram, as well as the video-sharing platform YouTube \citep{o2008user}, which can reflect human perceptions of environments as sensors and their contributions to the volunteered geographic information (VGI) \citep{goodchild2007citizens}. And affective memories are produced and archived in these technology-mediated platforms \citep{elwood2015technology}.
In those UGC, people express their emotions actively through tones of the voice \citep{schuller2009acoustic}, facial expressions \citep{ekman1993facial}, body gestures, and written forms \citep{bollen2011modeling}, or their emotions are captured passively by various types of sensors. Also, state-of-the-art AI technologies make it possible to collect human emotions from massive data sources and have revolutionized the research of human emotions. Several existing studies have tried to connect geography and collective emotions from those UGC using advanced technologies and got promising results \citep{mitchell2013geography}. However, the absence of attention to the role of place as locale to human emotions still exists \citep{smith2016emotion}. In addition, most existing research used natural language processing (NLP) to extract human emotions from textual corpus \citep{strapparava2004wordnet,cambria2012senticnet}. Such methods may face challenges such as multi-cultural differences in language, which may not be suitable for global-scale research (more discussions in Section \ref{S:literature} and Section \ref{S:discussion}). In comparison, the facial expression of emotions is said to be universal across countries and different periods, and can capture human emotions in real-time, which may be suitable for a place-based emotion extraction framework in a global-scale.

In this research, our goal is to investigate human emotions in places and explore potentially influential environmental factors. 
We term the study phenomenon as \textit{Place Emotion}, which is a special case of the general affective computing in geography, i.e., to examine the human emotions at different places with different affordances (including the environment and human activities). The research questions are as follows:
(1) How to extract and compute human emotion scores from amount of georeferenced photos taken in different places?
(2) What is the relationship between human emotions and environment factors at places?
To answer these questions, a general framework utilizing UGC to compute human emotion scores at places based on facial-expression recognition and spatial clustering techniques is proposed.
However, since there are many types of places with variety of environment factors, we only select one specific type of place (i.e., tourist attraction sites) as a case study to test the feasibility of our proposed workflow.

Tourist attractions, which attract ``non-local" travelers for sightseeing, participating activities, and experiences \citep{leiper1990tourist,lew1987framework}, are a popular type of places \citep{jones2008modelling}, and are located across the world, which are suitable for being chosen as a case study of global scale research.
In the past decades, with the growth of economy and the development of modern transportation techniques, tourism has experienced continued growth and deepening diversification to become one of the fastest growing economic sectors in the world \citep{ashley2007role}.
For a tourist, the choice of places to visit in planning a trip is the first step \citep{bieger2004information, sun2018building}, while the options are often numerous.
When retrieving information of tourist sites, a fair and comprehensive ranking list on tourist attractions is often useful.
However, existing ranking lists relying on the environment \citep{amelung2007implications} and socio-economic \citep{bojic2016scaling,chon1991tourism} aspects of the tourist sites.
These factors indeed influence travel flows but from an objective way.
The perceptions and feelings of tourists are often ignored. 
A ranking list based on human emotions might provide different insights from human-oriented preferences. Additionally, happiness is one of the most common basic emotions \citep{ekman1994nature,eimer2003role,izard2007basic}. Therefore, a ranking list of the happiest tourist sites in the world will be created as an outcome of the affective computing at each site.

To this end, this study presents a novel framework to measure human emotions at places from facial expressions and to explore influential factors to the degree of happiness at different places.
Tourist sites are taken as a specific type of place for experiment. 
The contributions of the study are three-fold. (1) We propose a novel approach for extracting and characterizing the average happiness score at each place using computer vision and spatial analysis techniques.  
(2) We explore the relationship between different kinds of environmental contexts and the degree of happiness extracted from human facial expressions.
(3) We create a ranking list of the happiest tourist sites based on crowdsourcing human emotions rather than objective indices, and provide new insights on integrating human emotions to enrich the understanding of sense of place in geography and in place-based GIS.

The remainder of paper is organized as follows. First, in the section \ref{S:literature} ``Related Work", we conduct the literature review on place emotion related studies. In the section \ref{S:method} ``Methodology", we present a methodology framework and explain our computational procedures. Then in the section \ref{S:experiment} ``Experiments and Results", we test the framework with a case study of human emotions at 80 worldwide tourist attractions. We discuss the implications and comparison of our image-based method to the text-based studies in the section \ref{S:discussion} ``Discussion''. Finally, we conclude this work and present our vision for future research in the section \ref{S:conclusion} ``Conclusion and Future Work".

\section{Related Work}
\label{S:literature}
There are two categories of affective computing.
One is about several instinctive basic emotions like happiness, sadness, anger, etc \citep{ekman1994nature}.
The other is to detect the polarity of sentiments like positive, neutral and negative expressions, which are organized feelings and mental attitude \citep{pang2008opinion}. Unless specifically clarified, we use the general term ``emotion" to represent both categories interchangeably in this paper. Both emotion and sentiment studies enable us to understand human perceptions of the society and the environment \citep{zeng2009survey}. Exploration and understanding of human emotions and sentiments have attracted volumes of interest from psychology \citep{ekman1993facial,berman2012interacting,svoray2018demonstrating}, biology \citep{darwin1998expression}, computer science \citep{lisetti1998affective}, geography \citep{davidson2004embodying,mitchell2013geography,svoray2018demonstrating,hu2019placesentiment}, and public health \citep{zheng2019airpollution}, just to name a few.

The emotion collection methods evolve over time. Traditionally, scholars from social sciences often use questionnaires and self-reports to investigate the emotions of people in different environmental contexts \citep{niedenthal2018heterogeneity}.
Several rankings of human's happiness are published in recent years, including the \textit{World Happiness Report} released by the United Nations Sustainable Development Solutions Network\footnote{\url{http://worldhappiness.report}}, which ranks the happiness of countries' citizens by investigating the social-economic indices.
The \textit{Measuring of National Well-being Program}, which is released by the Office of National Statistics, UK, monitors the well-being of citizens by producing assessment measures of the nation\footnote{\url{ https://www.ons.gov.uk/peoplepopulationandcommunity/wellbeing/articles/measuringnationalwellbeing/qualityoflifeintheuk2018}}.
The \textit{Gross National Happiness} is used in guiding the government of Bhutan with aspects of living standards, health, education, etc.
And the \textit{Satisfaction With Life Scale} measures the life satisfaction components of subjective well-being\footnote{\url{http://www.midss.org/content/satisfaction-life-scale-swl}}.
However, those methods encounter some challenges despite the spread usage in psychological science.
For example, it costs a lot of human resources and lacks timeliness \citep{golder2011diurnal}.
And the results relied on the questionnaires may have constraints of self-knowledge and psychological influence of informed consent \citep{baumeister2007psychology}.

With the emergence of affective computing technologies, more efficient ways for detecting human emotions are used.
Numerous studies on affective computing have been conducted and gained great success, especially using NLP methods to extract emotions from texts and explain from a geographic perspective. For example, \citet{mitchell2013geography} estimated human happiness at the state-level in the United States and explored the impact of socioeconomic attributes on human moods. \citet{ballatore2015extracting} utilized a corpus of about 100,000 travel blogs for extracting the emotional structure (including joy, anger, fear, sadness, etc.) of different place types.
\citet{bertrand2013sentiment} generated a sentiment map of New York city via extraction of emotions from tweet data.
\citet{zhen2018spatial} calculated the human emotion scores using the Weibo tweet data and explored the spatial distribution of sentiments in Nanjing. \citet{zheng2019airpollution} demonstrated that high levels of air pollution (e.g., PM 2.5) may contribute to the urban population’s reported low level of happiness in social media based on analytics of over 210 million geotagged tweets on Weibo.
 \citet{hu2019placesentiment} presented a semantic-specific sentiment analysis on online neighborhood textual reviews for understanding the perceptions of people toward their living environments.

Aside from the success, text-based measurements of emotions may encounter some challenges:
One problem is that texts are often recorded after events.
It means that the emotions expressed are not in real-time, but often after a period of transition.
The buffering time period may be beneficial to the user who expresses emotions. Because during a calm-down period, the user may utilize more dispassionate linguistic expression to maintain a stable social identity \citep{coleman2013feeling}.
Another challenge in extracting emotions from texts is the multi-lingual environment.
Different languages may vary in words and syntax for expressing the emotions.
Most of emotion extraction models are based on the words or the syntactic and semantic structures of the sentences, which are unique in each language \citep{shaheen2014emotion}.
So far, no existing method is established to standardize the emotional scores computed from all different language models.
Therefore, affective computing based on texts has been limited to analyze materials in one language at a time.
For problems being multi-lingual, text-based affective computing may be confronted with difficulties.

In comparison, image-based approaches \citep{zhang2018measuring}, especially facial expression-based emotion extraction methods have been improved greatly in recent years because of the emergence of deep convolutional neural networks \citep{yu2015image}, which even perform better than human in face-recognition benchmark testing \citep{wang2018deep}. \citet{svoray2018demonstrating} analyzed Flickr photos and found the positive relationship between human facial expressions of happiness and their exposure to nature with urban density, green vegetation, and proximity to water bodies in the city of Boston.
By extracting and identifying key points from face images based on facial activities and muscles, machine learning models can learn the visual patterns of faces according to the emotional labels \citep{calvo2010affect}. Therefore, the emotions of faces can be extracted. Each culture has its own verbal language, and emotion has its own language of facial expressions.
The relationship between emotions and facial expressions has been extensively explored.
\citet{levenson1990voluntary} pointed out that subjective emotions have significant connections to the facial activities, which provides the fundamental theories of facial expression-based affective computing. Facial expression-based emotion recognition methods have several advantages. First, facial expressions are both universal and culturally‐specific \citep{matsumoto1991cultural}.
Though connections between emotions and cultures varied \citep{cohn2007foundations}, strong evidence has been provided that there is a pan-cultural element in facial expressions of emotion \citep{ekman1970universal}. 
People from ancient times to the present, from all over the world, and even our primate relatives hold similar basic facial expressions, especially smiling and laughter \citep{preuschoft2000primate,parr2006understanding}.
It indicates that humans are universal when representing basic emotions and facial expression-based emotion extraction methods are suitable for global-scaled issues especially for solving the multi-lingual problem.
In fact, some existing researches have explored the worldwide expression of emotions based on facial expressions in photos \citep{kang2018happier}, which shows the universal compatibility of such methods.
In addition, facial expressions are produced spontaneously when emotions are elicited \citep{berenbaum1992relationship}.
By recording and analyzing facial expressions, researchers can track down emotions as they were just formed. As advanced computer-vision based systems and algorithms are becoming more mature, facial expressions as well as facial muscle actions can be recognized and computed with quantitative scores (possibilities) of recognized emotions \citep{ding2017facenet2expnet,kim2016hierarchical, zeng2009survey}. As a result, facial-expression based researches are spreading in affective computing.
For instance, \citet{kang2017mapping} examined the emotion expressed by users in Manhattan, New York City and compared the human emotions fluctuation with stock market movement to find out the relationship between the two.
\citet{abdullah2015collective} used images from Twitter to calculate emotions from facial expressions and compared them with socio-economic attributes.
\citet{singh2017individuals} analyzed the smiles and diversity via social media photos and pointed out people smile more in a diverse company.

In sum, considering that emotions can be recorded in real time and are universal in multi-lingual environment, facial expressions might be more suitable for place-based emotion extraction in a global scale, as places are located around the world with affording different groups of people and various kinds of activities. To the best of our knowledge, our research is among the pioneer studies that utilize the state-of-the-art facial expression recognition techniques and large-scale georeferenced photos for exploring the human emotions at different places at the global scale. 
\begin{figure}[H]
	\centering\includegraphics[width=0.67\linewidth]{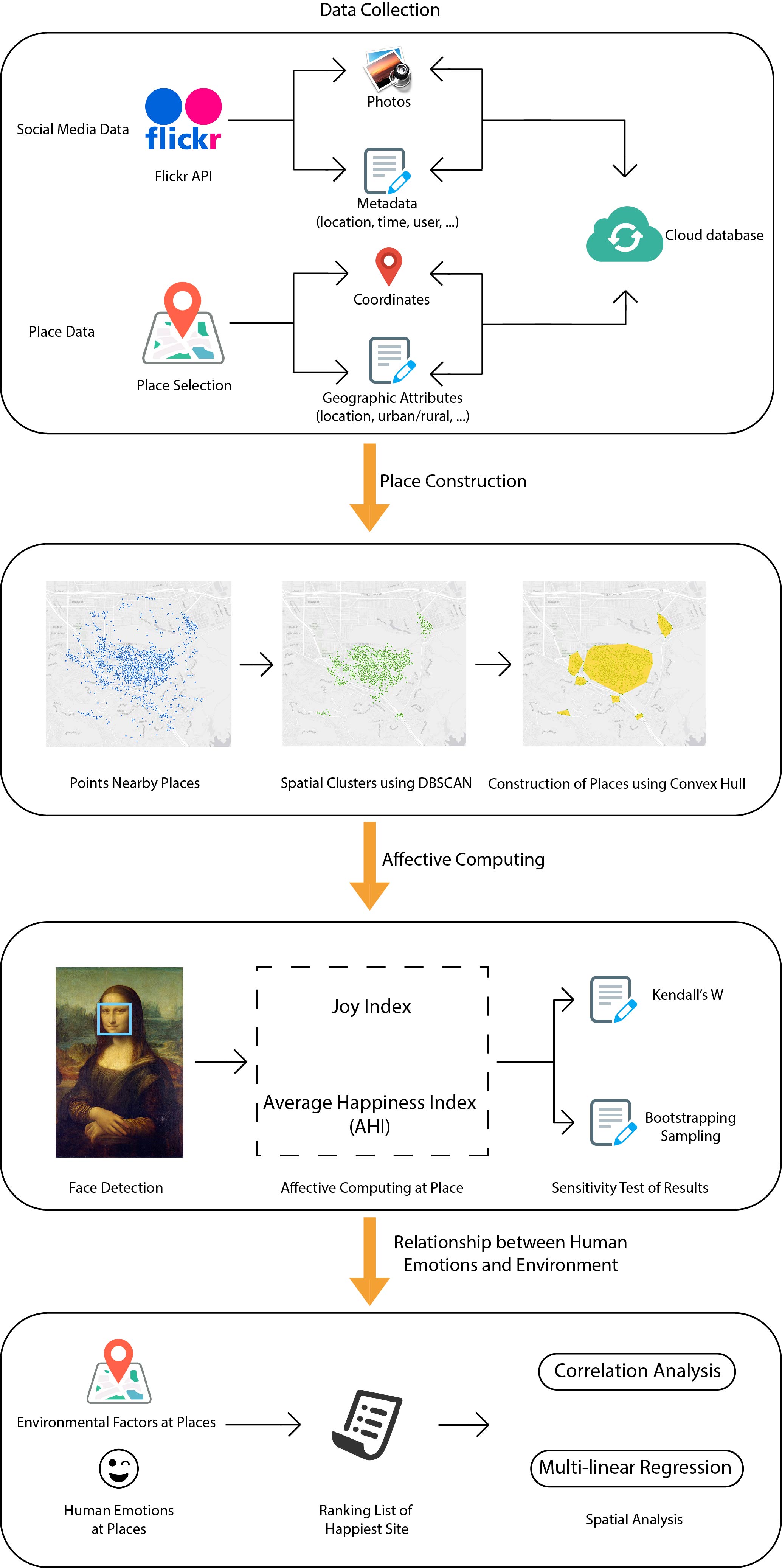}
	\caption{The workflow of this research.}
	\label{F:workflow}
\end{figure}

\section{Methodology}
\label{S:method}

\subsection{Framework}
\label{S:2.Framework}
As shown in Figure \ref{F:workflow}, there are four steps for the framework of extracting and measuring human emotions at different places.
First, large-scale geo-referenced crowdsourcing photos in social media are collected and positioned on the data server.
Several geographical and environmental attributes (e.g., the proximity to water bodies, openness, landscape type) of each place are also retrieved and recorded. 
Second, the footprints of “places” in our study are generated using the area of interest (AOI) extraction approach based on the spatial density of photos \citep{li2012constructing,hu2015extracting}.
Then, with the state-of-the-art cognitive recognition methods based on computer vision technologies (e.g., object detection and localization), human emotions are extracted and measured via facial expressions detected in the social media photos.
In order to examine whether results of affective computing are robust and solid, we also implemented sensitivity tests to check the concordance of results with varying algorithm parameter settings.

After the calculation of human emotions at different places, it is necessary to explore what environmental factors have influences on the expressions of human emotions.
Correlation analysis and multi-linear regression models are utilized to explore the relationship between human emotions and environmental factors.

\subsection{Data Preparation}
\label{S:2.Data}
There are two datasets used in this research.
One is the places as well as their geographic attributes for exploring the relationship between human emotions and environments.
And the other is georeferenced social media photos for affective computing, which are collected from the Flickr website based on the coordinate information of place names.

In many geographic information systems and digital gazetteers, places, are often represented as points of interest (POIs) although places have footprints that vary by type (e.g. points, lines, or polygons) \citep{goodchild2008introduction}.
Based on the place names, the coordinates of those place centers are harvested from the Google Maps Places Application Programming Interface (API)\footnote{https://developers.google.com/places/web-service/intro}. A list of geographic attributes and environment factors are recorded at each place (see section \ref{S:3.Data} for more details). 

Photos taken at different places are obtained from Yahoo Flickr platform.
Flickr is a publicly available social media platform where users can upload and share their photos, and it is one of the most commonly cited websites in the era of Web 2.0 \citep{cox2008flickr}.
Options for geo-tagging photos are also provided in the website as more and more GPS chips are embedded in smart phones and cameras. Time and geographical information are recorded automatically when saving photos from location-aware devices.
In addition, users can drag their photos on a map and input their locations for geo-tagging upon uploading. 
Therefore, each photo can be positioned on the map.
For most photos, the locations can be labeled correctly and the uncertainty of the data (eg. incorrect location of photo) can be removed by construction of the place introduced in Section \ref{S:2.Place0}.

Flickr’s API\footnote{https://www.flickr.com/services/api/} allows developers and researchers to collect a huge amount of data from the platform.
Public geo-tagged photos with information including user ID, photo ID, latitude, longitude, tag text, time stamp and so on, are retrieved and recorded within a certain distance to the center point at specific places.
And the center point coordinates of places are retrieved from the Google Place API.
Each photo is saved with its original resolution while keeping a link pointing to its original URL.
All the information is stored in a database for data manipulation.

\subsection{Construction of Places}
\label{S:2.Place0}
As place is a product of human conceptualization that is derived from human experience to describe a specific space \citep{tuan1977space, couclelis1992location, curry1996work,merschdorf2018revisiting}, one main challenge for modeling places in GIS is the vague boundaries \citep{burrough1996geographic,montello2014vague, gao2017data}. The boundary is often generated from the density estimation and spatial clustering of georeferenced photos \citep{feick2015multi, hu2015extracting}. 
In this reseasrch, places are constructed by the following steps based on user generated photo footprints:
(1) utilizing a density-based spatial clustering for application with noise (DBSCAN) to extract the hotspot zones of human activities.
(2) using the convex hull to find out the minimum bounding geometry based on a set of points remained after spatial clustering.

DBSCAN, which is a point-based spatial clustering algorithm \citep{ester1996density}, is used to identify clusters of geo-tagged photos. Compared with the K-means clustering algorithm, DBSCAN can find arbitrarily shaped clusters, and it does not require the predefined number of clusters in advance. 
In addition, it is relatively stable and robust to the noise data. Some geo-tagged photos are manually uploaded by users without specific criteria and may generate noise; for example, a user may drag photos to wrong locations.
By applying the DBSCAN algorithm, those noise data can be removed and the core areas of each place, in other words, the hotspots where users most likely to stay and to take photos, will be remained. 

The DBSCAN algorithm requires two parameters, namely \(\varepsilon\) and \(minPts\). The
\(\varepsilon\) is the search radius representing the maximum distance of the search neighborhood to the center point. And the \(minPts\) indicates the minimum number of points a cluster should have. Different settings of the two parameters will influence the result, and proper values according to the characteristics of places should be selected.
As suggested by several previous research \citep{hu2015extracting,mai2018adcn,liu2019exploring}, a value between 40 m to 300 m of \(\varepsilon\) is suggested in clustering human activities.
As the number of photos and users may vary in different places, it is not suitable to use an universal absolute number as \(minPts\). 
Therefore, a percentage of the number of photos at a certain place is used as \(minPts\).
Consequently, a combination of different parameter settings should be tested to find out the best parameter combination.

After the spatial clustering of photo locations, the next step is to derive the core areas of places from the clustered points.
The convex hull is a high-quality geometric approximation method used for efficiently clustering geographical features \citep{graham1972efficient,barber1996quickhull,liu2019exploring,yu2014object}.
A convex hull is the minimum bounding polygon containing a set of points. It has been utilized in a number of studies to find the minimum bounding shape of the clustered points \citep{liu2019exploring}.

\begin{figure}[H]
   \centering\includegraphics[width=1.0\linewidth]{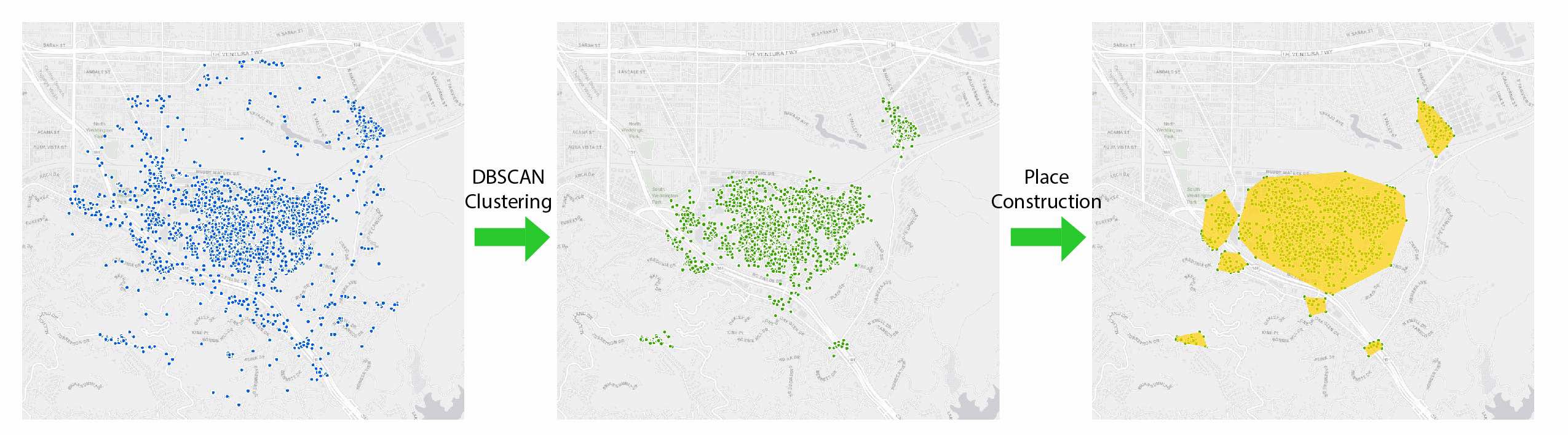}
   \caption{Construction of a place based on spatial clustering and the convex hull approach.}
   \label{F:place}
\end{figure}

Figure \ref{F:place} shows the process of construction of places that are represented as polygons generated from the aforementioned steps.

\subsection{Measurement of Human Emotion}
\label{S:2.Emotion}
One main research question in this work is how to extract basic human emotions and to quantify the degree of happiness expressed by users at different places. The state-of-the-art computer vision and cognitive recognition technologies make it possible to extract and quantify human emotions from facial expressions.
In this study, we propose two indices, namely the ``Joy Index" and the  ``Average Happiness Index", to measure the degree of ``happiness atmosphere" at each place.

\subsubsection{Affective Computing}
We used the Face++ Emotion API\footnote{https://www.faceplusplus.com/emotion-recognition/} to detect human faces in photos and to extract human emotions based on their facial expressions. 
The Face++ platform is a mature commercial cloud-computing enabled AI technology provider with a great number of customers and developers using its products, and is said to perform well in several facial recognition-related competitions\footnote{https://www.faceplusplus.com/blog/article/coco-mapillary-eccv-2018/}, which proves the reliability of the system, and hence is selected for extracting emotions from human faces.
A set of computer vision-based services are provided for human facial recognition and analyses.
The attributes of all faces in a photo are extracted, including the face position and extent, human emotion, age, ethnicity, gender, and even beauty. The Face++ API produces two measurements for evaluating the emotion of human faces.
One is the \textit{smile}, which describes the smile intensity \citep{whitehill2009toward} and includes two elements \(smile\ value\) and \(smile\ threshold\).
The \(smile\ value\) is a numeric score (from 0 to 100) to indicate the degree of smiling while the \(smile\ threshold\) is provided by the cloud AI system to judge whether the detected face is smiling or not.
Generally, if the \(smile\ value\) is larger than the \(smile\ threshold\), the face is judged as a smiling face.
Therefore, based on the \textit{smile} attribute, each face in the photo is classified as either smiling or not-smiling.
The other measurement is \textit{emotion structure}, which is a vector of scores (from 0 to 100) to describe seven basic emotional fields:
anger, disgust, fear, happiness, neutral, sadness and surprise.
All scores of one face sum up to 100.
The higher the score is, the more confidence an emotion represents.
Hence, the emotion field could illustrate the intensity of a particular emotion from different dimensions.

\begin{figure}[H]
   \centering\includegraphics[width=0.95\linewidth]{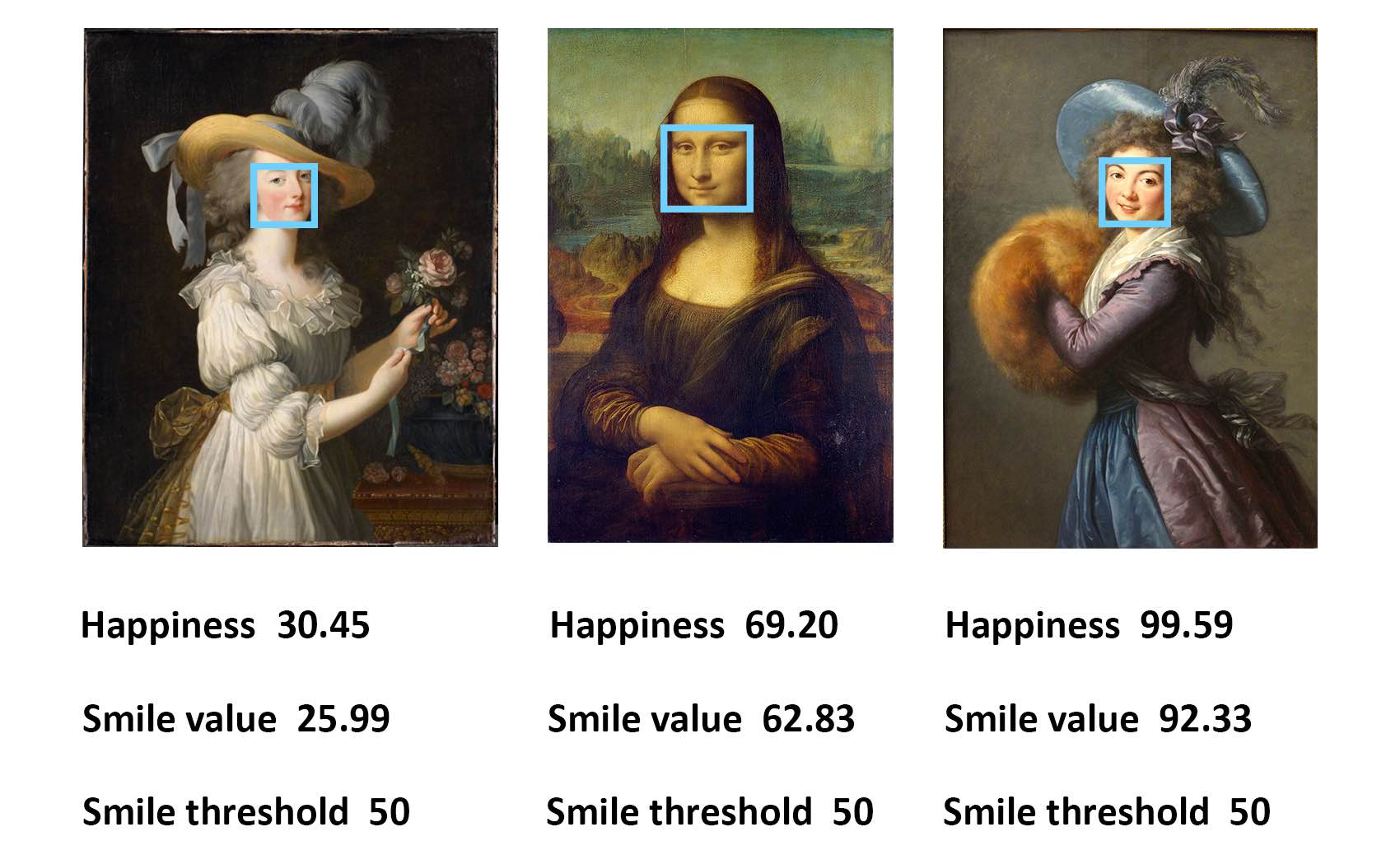}
   \caption{Emotional indices calculated for faces. (Source: Face++)}
   \label{F:emotion_api}
\end{figure}

It is worth noting that not all emotion fields are used in this study. 
Happiness is often recognized as one of the most common basic emotions \citep{izard2007basic}.
Although some arguments exist \citep{frank1993not}, smile can represent happiness in general. In addition, as happiness is the clearest emotional domain compared with all other dimensions of emotions \citep{wilhelm2014test}, we only use happiness value from the \textit{emotion structure}. Figure \ref{F:emotion_api} shows the happiness scores extracted from different human faces in photos. Notice that the actual human faces rather than those on paintings will be detected and analyzed in the experiments. 

\subsubsection{Emotional Indices for Places}
Two place-based human emotion measurement indices are proposed to evaluate the degree of happiness at different places in this study. Namely, the ``Joy Index" based on the smiling score and the ``Average Happiness Index" based on the happiness score.

The ``Joy Index" is calculated with consideration of the normalized difference between the number of smiling faces and the number of non-smiling faces using geo-tagged photos within the spatial footprint of each place as follows.
\begin{equation}
   J_i = \frac{C_s-C_{ns}}{C_s+C_{ns}}
\end{equation}
Where \(J_i\) is the joy index calculated at place \(i\), \(C_s\) is the number of smiling faces in the photos within this place, \(C_{ns}\) is the number of non-smiling faces.
The range of this index is between -1 to 1, a symmetric closed interval. 
A positive value represents that more people are smiling at a place, which indicates positive emotion conditions. While a negative value represents that more people don't have smiling faces, which may indicate a serious atmosphere at that place.

In comparison, the ``Average Happiness Index (AHI)'' calculates the average of happiness values for all detected faces in those geo-tagged photos at a place.
\begin{equation}
   AHI_i = \frac{1}{n} \sum_{j=1}^{n} H_j(i)
\end{equation}
Where  \(H_j(i)\) is the happiness value of human face $j$ at place \(i\).
The AHI illustrates the average degree of happiness for people at each place.

\subsection{Sensitivity Tests}

\subsubsection{Test for Place Construction}
During the construction of place described in section \ref{S:2.Place0}, a set of combinations for parameters \(\varepsilon\) and \(minPts\) are used.
Although the shape of place boundaries may vary with different parameters, the derived place emotion results should have a similar distribution and trend if the proposed approach is stable.
In order to check this, human emotion scores are calculated at each place with different parameter settings.
Then, the Kendall’s coefficient of concordance (W) is utilized to measure the agreement among those different human emotion detection results.

In order to do so, a ranking of places based on the detected average happiness score is created for each pair of \(\varepsilon\) and \(minPts\).
Assuming there are \(m\) combinations of parameters for \(n\) places.
Summing up the ranks $r_{i,j}$ in all $m$ scenarios in which a place \(i\) gets a total rank \(R_i\) via Equation \ref{E:R_rank}.
Then, \(\overline{R}\) is calculated as the average value of those total ranks across all places via Equation \ref{E:R_overline}, and the sum of the squared deviation \(S\) is calculated via Equation \ref{E:S}.
And the Kendall’s W can be calculated by Equation \ref{E:W}.

\begin{equation}
R_i = \sum_{j=1}^{m} r_{i,j}
\label{E:R_rank}
\end{equation}

\begin{equation}
   \overline{R} = \frac{1}{n} \sum_{i=1}^{n} R_i
   \label{E:R_overline}
\end{equation}

\begin{equation}
   S = \sum_{i=1}^{n} (R_i - \overline{R})^2
   \label{E:S}
\end{equation}

\begin{equation}
   W = \frac{12S}{m^2(n^3 - n)}
   \label{E:W}
\end{equation}

In general, if the test statistic \(W\) is 1, it means all judges with different parameter settings are assigned as the same order of the places. 
While \(W = 0\) indicates that there is no agreement among all judges and the ranks are in random.
If the results of W prove that the emotion score rankings are similar with different parameters in place construction, the influence of shape during the place construction process is limited. 
Also it would also show that the emotion scores calculated at each place are solid.

\subsubsection{Test for Affective Computing}
As the number of photos varied at different places, it is necessary to know whether the data collected are sufficient for human emotion calculation.
To test the reliability and stability of the facial-expression based on emotion recognition results, a bootstrapping strategy was applied to assess the robustness of the emotional indices calculated in the section \ref{S:2.Emotion}. 

Bootstrapping is a resampling approach proposed by Efron \citep{efron1992bootstrap}.
It is often used to approximate the distribution of the test samples.
By doing this, a confidence interval can be derived to show how confident the range of emotion scores is at each place.
The step-by-step details are described as follows:

\begin{enumerate}[label={(\arabic*)}]
   \item Assume that \(n\) faces are collected at place \(i\) as a sample set \(D'\). 
   Perform \(n\) times random sampling with replacement to form a new sample set with the same size as \(D\). Note that more than one face may exist in \(D\).
   \item Then,  the emotion indices \(e\) of the new sample set \(D\) is calculated.
   \item Repeat the two steps above for \(N\) times to generate \(D_1, D_2, ..., D_N\) with emotion results \(e_1, e_2, ..., e_N\).
   \item Rank the affective computing results and calculate the average value of the emotional indices as the final output of the place \(i\).
   Discard the lowest 2.5\% and the highest 2.5\% results.
   The remaining results show the 95\% confidence interval of emotional indices calculated at the place \(i\).
\end{enumerate}

The results of bootstrapping show the confidence intervals of the possible emotion scores at each place, which help evaluate the stability of the emotion calculation results.
Although it is impossible to know the true confidence interval as photos are collected in bias anyway, the derived results are more asymptotical to be the truth \citep{diciccio1996bootstrap}.
Further analyses are conducted based on the emotional results after the bootstrapping processing.

\subsection{Influence of Environment Factors}
\label{S:2.Environment}
As suggested by environmental psychology studies \citep{capaldi2014relationship,svoray2018demonstrating}, human emotions can be affected by the surrounding environment. 
Therefore, exploring the potentially influential geographical and environmental factors and their importance has great significance to understand human emotions at different places.
To do so, the Pearson's correlation analysis \citep{benesty2009pearson} and the multiple linear regression (MLR) were employed in this study.

As mentioned in section \ref{S:2.Data}, a group of social and physical geographic attributes are collected when retrieving the information for each place. Those factors are represented as \(a_1, a_2, ..., a_n\) at each place. Please note that since this paper aims at proposing a general computational framework for extracting place emotions, and the environmental factors may vary in different types of places.
Therefore, we do not define a complete set of factors in this research and further research is needed for enumerating a complete list of variables related to a specific type of place.
As a case study, by referring several existing works and our geographical knowledge, several environmental factors are chosen in this work as described in Section \ref{S:3.Data}.

The Pearson's correlation coefficient $\rho$ is employed to explore the positive and negative impacts, and the strength of linear relationship between an environmental factor and the emotion score at each place. As correlation analysis is only suitable for numeric values, for categorical variables (e.g., continents), the correlation coefficient between the emotion and each category is calculated by converting categorical variables to dummy variables (0,1).

For each influential factor \(a\), the correlation analysis was performed with a combination of one emotion index \(e\) via equation \ref{E:Correlation}:
\begin{equation}
   \rho_{e, a} = \frac{E[(e-\mu_e)(a-\mu_a)]}{\rho_e \rho_a}
   \label{E:Correlation}
\end{equation}

Where the Pearson's correlation coefficient \(\rho_{e,a}\) is calculated with the expected covariance value \(E\) of the two variables \textit{e} and \textit{a} with their mean values \(\mu_e\) and \(\mu_a\), and the standard deviations \(\rho_e\) and \(\rho_a\). 
A positive value shows that the factor has positive impacts on the emotion index \textit{e} and vice versa.

And the MLR uses all geographical and environmental variables \((a_1, a_2, ..., a_n)\) to predict the emotion index value $E_i$ at each place \(i\) as:

\begin{equation}
   E_i = f(a_1 + a_2 + ... + a_n) + \gamma
   \label{E:Ei1}
\end{equation}

Where \(\gamma\) is an unobserved error term. The impact of each attribute could be measured using the coefficient of each independent variable. The \(R^2\) is calculated as a goodness-of-fit statistic to determine how well the MLR model fits the observed place emotion data.
\section{Experiments and Results}
\label{S:experiment}
\
As the experience of travel and tourism is deeply connected with the place \citep{wearing2009tourist}, we take tourist attractions as a case study place type to examine the feasibility of our \textit{Place Emotion} sensing framework.

\subsection{Input Datasets}
\label{S:3.Data}
Tourist sites selected in this study are located around the world.
And the sites selected have to be famous regarding the annual number of visitors, comprehensive in terms of cultural representativeness, and diverse in terms of site types.
In addition, in order to get reliable emotion detection results, the site should have a large number of photos taken and uploaded by tourists.
To find them, several official resources\footnote{https://whc.unesco.org/en/map/} and open statistics were checked\footnote{https://www.lovehomeswap.com/blog/latest-news/the-50-most-visited-tourist-attractions-in-the-world}\(^,\)\footnote{https://www.travelandleisure.com/slideshows/worlds-most-visited-tourist-attractions}.
Those selected attractions are distributed all over the world and listed in Figure \ref{F:map}.
In total, there are 80 sites, including 24 sites located in Asia, 25 in Europe, 29 in North America, and only 1 site located in Africa and in Oceania respectively.
They are from 22 countries around the world.
The spatial distribution of all these tourist sites can be found in the Figure \ref{F:map}.

A group of geographical attributes and environmental factors are searched and recorded, to the best of our knowledge, which may influence the tourists’ degree of happiness at each site.
As human emotions are complex and influenced by multiple individual and environment variables, we only selected a small group of variables according to some existing studies from environmental psychology \citep{white2010blue,capaldi2014relationship,svoray2018demonstrating}. Other socio-economic and environmental factors as well as individual differences may be added in future work to explore. The selected variables are:
\begin{enumerate}[label={(\arabic*)}]
	\item The coordinates of the site location which are searched by the Google Maps Place API.
	\item The continent where the site is located.
	\item The country where the site is located.
	\item The existence of water bodies.
	As suggested by several related studies from psychology, landscape containing water bodies can influence human activities, and consequently affect moods \citep{white2010blue}. Therefore, taking water bodies into consideration is necessary.
	There are two circumstances where the water bodies exist. 
	One is that the water body exists within the tourist place. 
	The other is that the water body is nearby the tourist site and can be directly viewed by the people.	Otherwise, we deem that there exists no water bodies at that place.
	\item The distance to the nearest water bodies, which calculates the shortest distance from the nearest water bodies (lakes, oceans, etc.) to the place.
	If the water body exists within the site, the distance is 0.
	\item Whether the main part of the site is in an open or closed space.
	Most of previous studies have proved that activities in an outdoor environment have a positive effect on happiness \citep{thompson2011does}.
	Hence, the tourists sites are classified as open or closed space.
	Parks, squares, lakes, etc., which are open to the air,  are defined as open space, while sites like museums, stations, cathedrals, etc., whose main contents are indoor, are considered as closed space.
	\item The green vegetation coverage of each place.
	Several studies suggested that green space could reduce pressure and has positive impact on mental health \citep{maas2009morbidity,thompson2012more}. In order to measure the green space and its impact on the human emotions, the Normalized Difference Vegetation Index (NDVI), which is widely used in remote sensing of vegetation \citep{goward1991normalized}, was harvested from NASA Earth Observations\footnote{https://neo.sci.gsfc.nasa.gov/view.php?datasetId=MOD\_NDVI\_M}. The NDVI product in June 2017 was downloaded and values were spatially joined to each site.
	\item Whether the place is located at an urban or rural environment.
	Similar to the open or closed space, urban areas which have higher building density than rural areas which have more natural environment, have great influence on human emotions \citep{wooller2018can}.
	
	\item The type of a tourist site.
	Different types of tourist sites may have different groups of visitors and mobility patterns. The type of tourist sites may be associated with the mental conditions \citep{leiper1990tourist}.
	Based on the site type defined by the Google Maps Place API as well as several tourism-related studies \citep{lew1987framework}, there are six types of tourist attractions defined in this research. Namely, \textit{natural} (like waterfalls, where places have limited human-made things), \textit{amusement} (like the Disneyland theme parks, where tourists visit places for enjoying games and other activities for fun), \textit{religious} (like cathedral, where people visit mostly for religious-related activities), \textit{museum} (like the Metropolitan Museum of Art, where historical, scientific, and artistic objects are kept), \textit{palace} (like the Forbidden City, where old cities and castles located), and \textit{other cultural categories} (like the Grand Bazaar, where places have cultural and historical values, but not belong to other categories aforementioned). Note that the six types are selected only based on the attributes of the 80 tourist attractions. More types of tourist attractions can be defined in other datasets.

\end{enumerate}

After the selection of tourist attractions, all photos taken between Jan. 2012 and Jun. 2017, within the distance of 1km to the center of each attraction site were downloaded from the Flickr website.
The search radius is larger than the spatial footprint of a place in most cases to ensure the number of photos harvested is sufficient.
In total, 6,199,615 photos were collected.

\subsection{Construction of Place and Affective Computing}
\label{S:3.Method}
Following the steps in section \ref{S:2.Place0} and \ref{S:2.Emotion}, each tourist site is constructed by the user generated footprints with the DBSCAN spatial clustering and the convex-hull minimum bounding geometry algorithms, and the place emotions are calculated. In total, 2,416,191 faces are detected and evaluated, and the ratio between the number of faces to all the photos is about 38.97\% while the  proportion of pictures with faces is about 20\%. For each site, two emotional indices, namely the Joy Index and the Average Happiness Index (AHI) are calculated by the faces remaining within each site.
However, since different parameters settings of DBSCAN in the place construction process may impact the generated sites, a set of combinations of parameters are tested. In the experiment, we iteratively chose the \(\varepsilon\) as 50 m, 100 m, 200 m, 300 m, and the \(minPts\) as 0.5\%, 1\% and 2\% according to the recommendations of previous studies \citep{hu2015extracting,gao2017data}. In sum, 12 combinations of parameter settings were experimented individually and applied into the \textit{Kendall's} concordance test.

For each pair of parameters, a ranking of sites are returned based on each emotional index.
The output of \textit{Kendall's W}  is 0.99 for 12 rankings based on the normalized Joy Index, and the same value 0.99 for all rankings based on the AHI, and 0.98 for 24 ranking lists including both indices. The results illustrate that all pairs of parameters can result in a very similar ranking order of the happiest places, which means that the proposed method is stable and the selected parameters of DBSCAN have limited impact on the overall place emotion ranking. The happiness indices calculated at each place are almost invariant in all the different experiments.
According to this, we chose only one parameter setting: 100 m as \(\varepsilon\) and 1\% as \(minPts\) for further analyses.

\begin{figure}[H]
	\centering\includegraphics[width=0.7\linewidth]{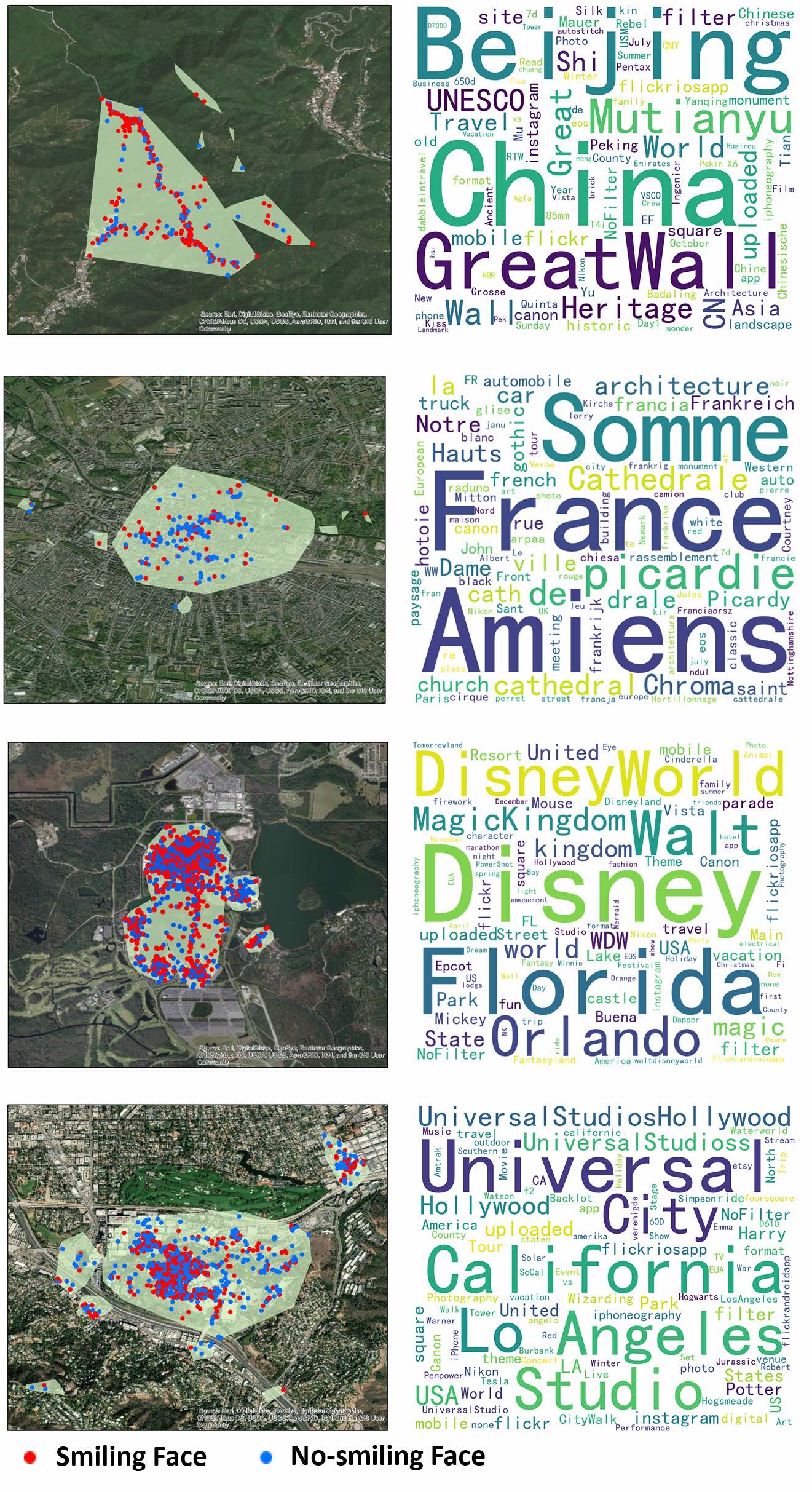}
	\caption{The spatial distribution of Flickr photos with smiling faces and without smiling faces and their most frequent word tags across four sample tourist sites: the Great Wall, the Amiens Cathedral, the Magic Kingdom Park at Disneyland, and the Universal Studio Hollywood.}
	\label{F:case_study} 
\end{figure}

As for the exploration analysis, four famous tourist attractions of interest: the Great Wall, the Amiens Cathedral, the Magic Kingdom Park at Disneyland, and the Universal Studio Hollywood, are selected as individual examples to demonstrate the specific place emotion distributions (in Figure \ref{F:case_study}). 
The first column of figures show the spatial distribution of photos with smiling faces and without smiling faces inside the place constructed, while the second column of the figures show the most frequent word tags shared by the Flickr users across those sites. The constructed places are multi-part polygons and the photos outside the polygons are removed in order to reduce the data noise. The red points show smiling faces while blue points indicate no-smiling faces. 
According to the figure, the Great Wall, the Magic Kingdom Park at Disneyland and the Universal Studio Hollywood have more smiling tourists while tourists at the Amiens Cathedral have less smiling faces as people in the religious site may be less inclined to smile. Moreover, the semantics of the photos are also explored. The word cloud visualization shows the top 100 tags of those social media photos at each place.
In addition to the country, department (in France) and city names, a list of tourist site names including Mutianyu, Great Wall, Cathedrale, Disney World, Universal Studios are identified from those geotags, which indicate that those photos could represent the place information although not all the words and topics are necessarily indicative to a specific place \citep{adams2012geo,adams2013inferring}. These examples show that the computational framework for emotion extraction based on facial-expressions at places is generally effective.

\subsection{The World Ranking List of Happiest Tourist Attractions}
\label{S:3.Ranking}
\begin{figure}[H]
	\centering\includegraphics[height=0.74\textheight,angle=90] {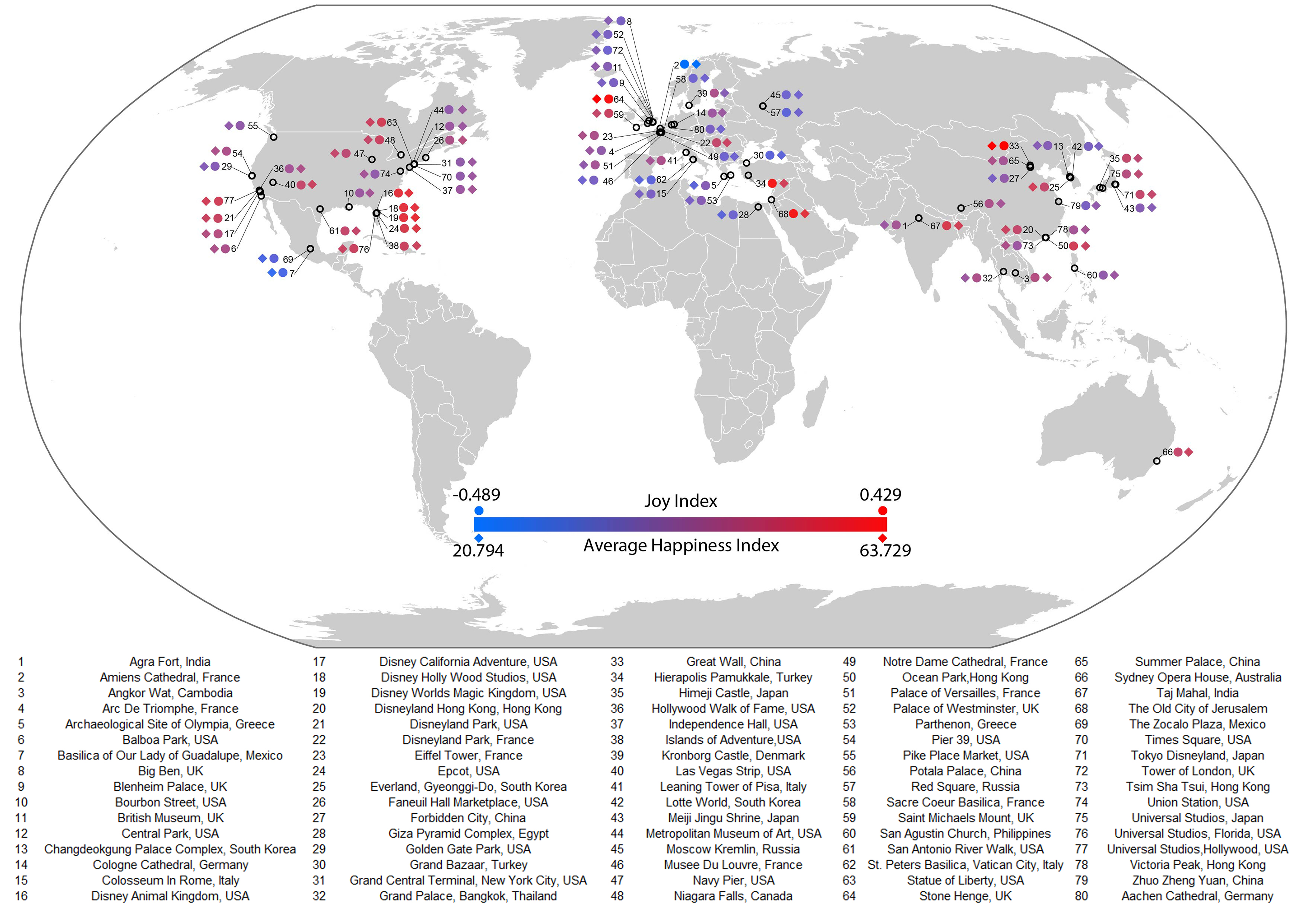}
	\caption{The spatial distribution of all tourist sites and their associated happiness indices.}
	\label{F:map}
\end{figure}

\begin{figure}[H]
	\centering\includegraphics[width=1.0\linewidth]{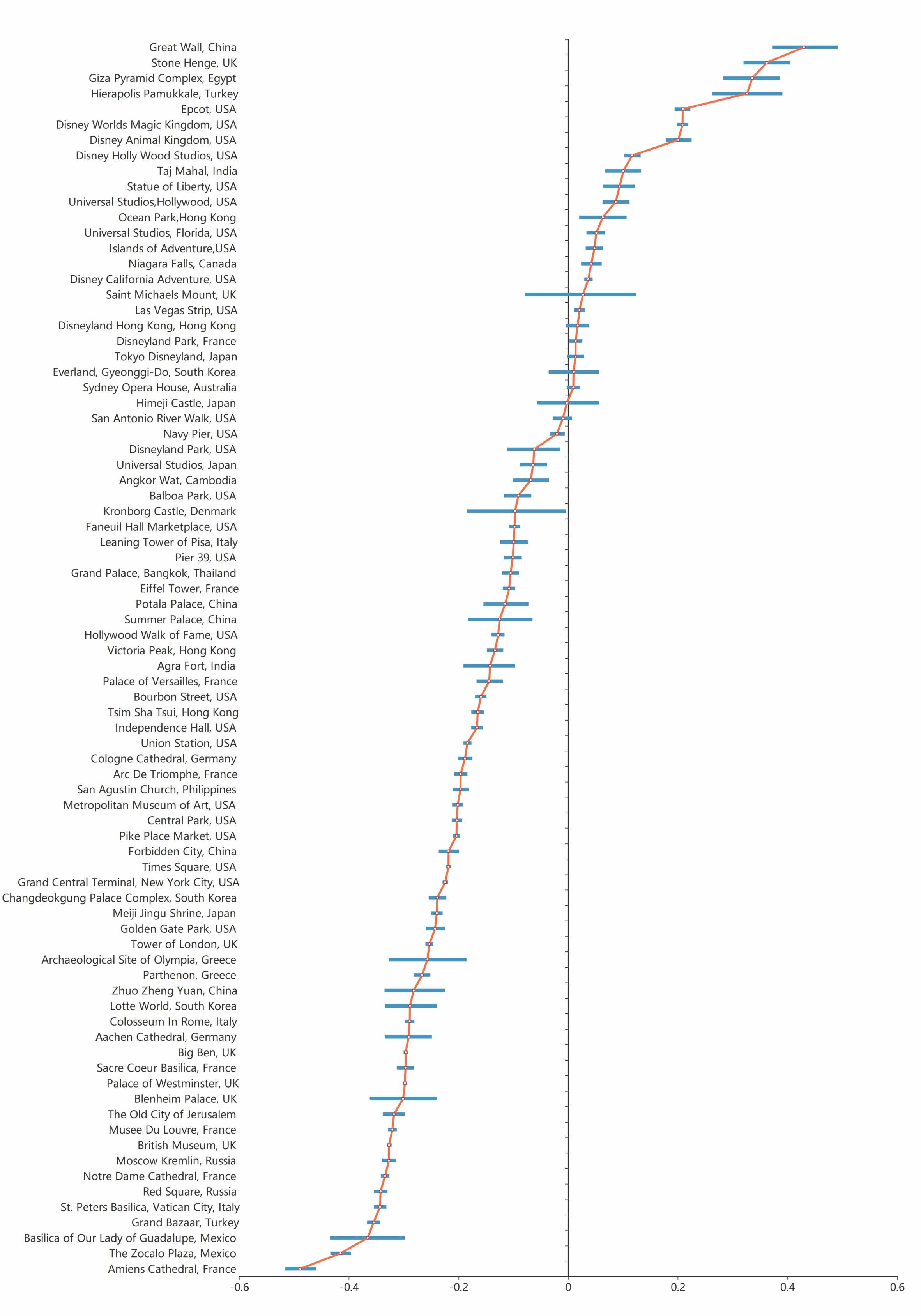}
	\caption{The ranking list of tourist sites based on the Joy Index. The 95\% confidence interval of the emotional index at each site is characterized by the blue bars, and the circles indicate the averaged values of the emotional index. (Note: figure is zoomable)}
	\label{F:ranking_joy} 
\end{figure}

\begin{figure}[H]
	\centering\includegraphics[width=1.0\linewidth]{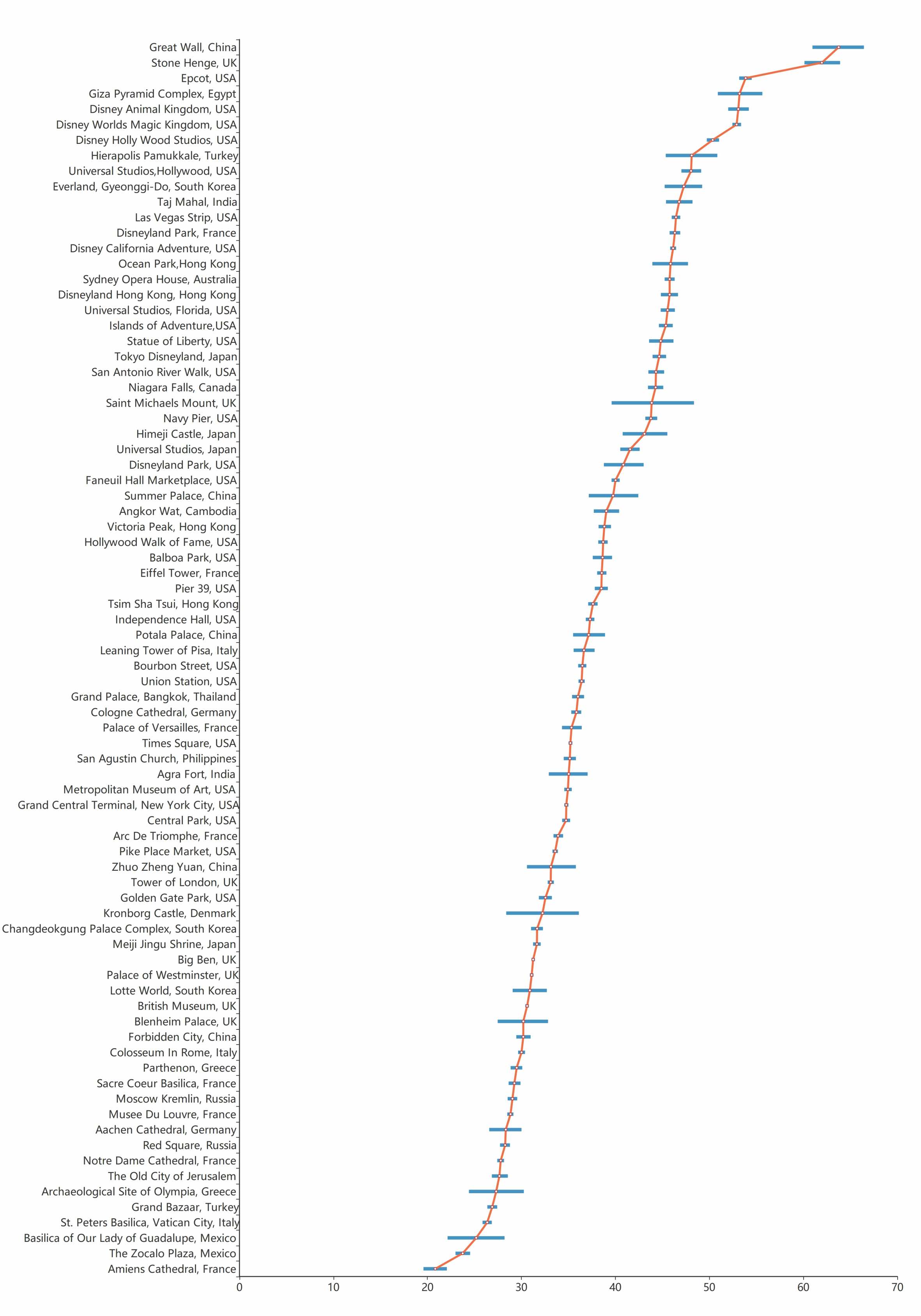}
	\caption{The ranking list of tourist sites based on the Average Happiness Index. The 95\% confidence interval of the emotional index at each site is characterized by the blue bars, and the circles indicate the averaged values of the emotional index. (Note: figure is zoomable)}
	\label{F:ranking_happiness} 
\end{figure}

Figure \ref{F:map} shows the spatial distribution of tourist sites as well as their emotional indices. 
The circles represent the joy index while diamonds represent the AHI of each site.
A deeper red color shows more happiness at a site while a deeper blue indicates less happiness. For each site on the map, its associated name can be found via the index.
Based on the emotional indices, two ranking lists of tourist attractions were generated in Figure \ref{F:ranking_joy} (Joy Index) and in Figure \ref{F:ranking_happiness} (AHI).
After using the bootstrapping strategy, the 95\% confidence interval of the emotional indices at each site is characterized by the blue bars, and the circles at the center of the lines indicate the averaged values of emotional indices.
For the joy index, a positive value represents more enjoyment smiling faces indicating a happiness atmosphere, while a negative value indicates that happiness cannot be deduced clearly from facial expressions in a site.
The average joy index across all sites is about -0.115, a little bit lower than 0, while the average of all AHI values is about 38.04.
The correlation analysis result shows that the Pearson's correlation coefficient between the two rankings is 0.97, which means that the two rankings are similar. 
Interestingly, the official slogan for Disneyland is ``The Happiest Place On Earth". However, according to the ranking lists from user generated crowdsourcing data, the top site that has the highest happiness indices in the world is the Great Wall, China based on the two measurements (Joy Index: 0.429, and AHI: 63.72).
But several amusement parks like the Disneyland Parks, the Everland in South Korea, the Ocean Park in Hong Kong indeed appear with high rankings though, which is in accordance with the public opinions. Meanwhile at the bottom of the ranking list is the Amiens Cathedral, with only -0.489 for the joy index and 20.79 for AHI. It is worth noting that low happiness scores don't necessarily mean that people at those sites (e.g., religious places) are not as happy as people in other types of places, but it could mean that people are less inclined to smile at those sites. However, since only tourist sites are chosen in our case study, most smiling faces are enjoyment smiles and seem to be associated with positive emotion and happiness in the top ranked sites.

\subsection{Relationships between Human Emotions and Environmental Factors}
\label{S:3.Relationship}

\begin{figure}[H]
	\centering\includegraphics[width=0.8\linewidth]{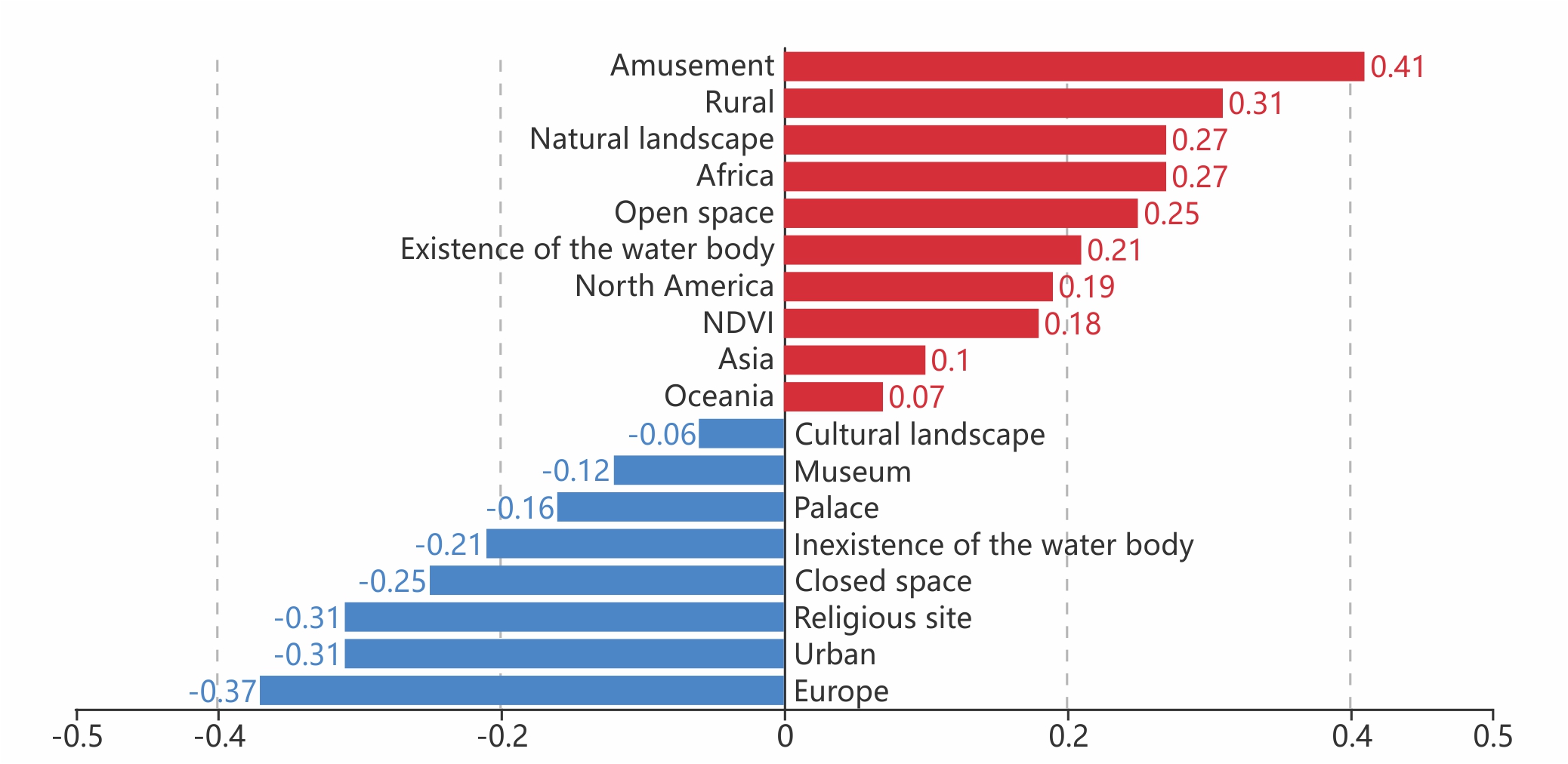}
	\\(a)\\
	\vspace{2em}
	\centering\includegraphics[width=0.8\linewidth]{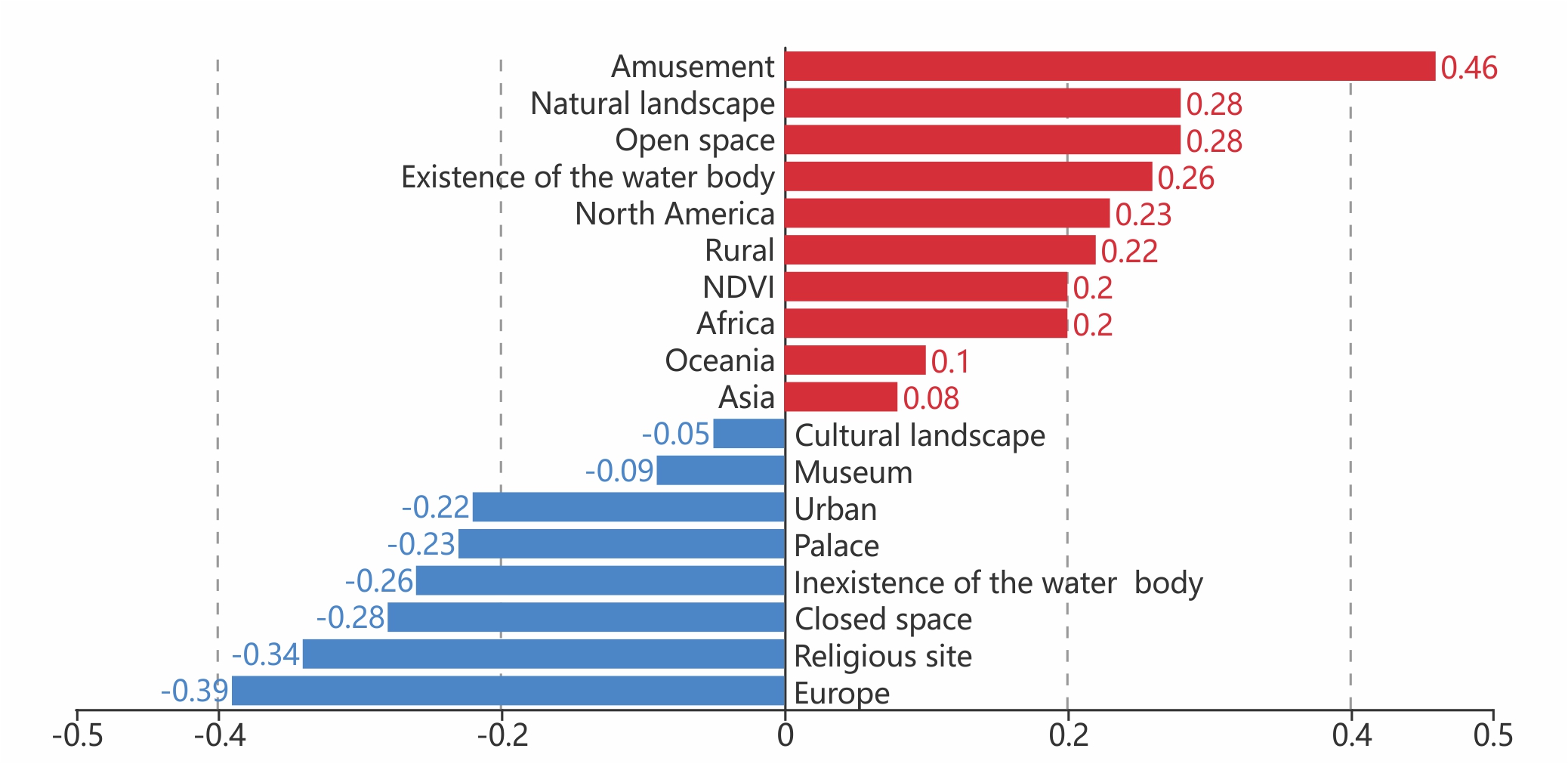}
     \\(b)
	\caption{The Pearson's correlation coefficients of the geographical and environmental attributes to the human emotions: (a) Joy Index; (b) Average Happiness Index}
	\label{F:correlation_joy_ahi}
\end{figure}

Each tourist site listed in Figure \ref{F:map} was assigned with a set of aforementioned attributes, namely, the continent, open or closed space, urban or rural area, attraction type, vegetation coverage, water body existence, and the distance to the nearest water body.
Figure \ref{F:correlation_joy_ahi}, Table \ref{T:multilinear_joyindex},  and Table \ref{T:multilinear_ahi} show the results of correlation analysis and multiple linear regression of the emotional indices and those attributes.

Results of the correlation analysis show that amusement parks have significant positive impact (0.41 in Joy Index and 0.46 in AHI) on the tourists' smiles and happiness, which is in accordance with our common knowledge.
As tourists often go to amusement parks to relax and enjoy holidays, they may be happier than going to other places.
Natural landscapes (0.27 in Joy Index and 0.28 in AHI), open space (0.25 in Joy Index and 0.28 in AHI), existence of the water body (0.21 in Joy Index and 0.25 in AHI), North America (0.19 in Joy Index and 0.23 in AHI), rural areas (0.31 in Joy Index and 0.22 in AHI), vegetation coverage by NDVI (0.18 in Joy Index and 0.2 in AHI) all have positive impact respectively.
Except for the continent variable, the coefficients of those aforementioned variables hint that, to some degree, places with more open environment can increase the degree of happiness of tourists with more enjoyment smiles.
On the contrary, compared with sites in other continents, people staying in the sites in Europe (sites selected in our case study only) may not explicitly express as much happiness with smile as that in other continents. What is more, religious site (-0.31 in Joy Index and -0.34 in AHI), closed space (-0.25 in Joy Index and -0.28 in AHI), nonexistence of water body (-0.21 in Joy Index and -0.26 in AHI), Palace (-0.16 in Joy Index and -0.23 in AHI) and urban areas (-0.31 in Joy Index and -0.22 in AHI) have negative impacts on the average happiness score of tourists.

According to the MLR results (Table \ref{T:multilinear_joyindex} and Table \ref{T:multilinear_ahi}), the impact of most variables show similar results with the correlation analysis.
The impact of Europe on the happiness conditions is negative and is statistically significant in the regression model. Sites with water bodies have a positive impact on human happiness but are not significant in our samples. Conversely, for sites located in urban areas, the emotional indices are negative and this result is significant. Natural landscape has positive impacts on the happiness indices but is not statistically significant.
The goodness of fit \(R^2\) is about 0.57 and statistically significant with p-value $<$ 0.001 for both indices, showing that the variation of human emotions at different places can be explained by those geographical and environmental factors to a certain degree.

In addition, as the type of tourist sites has impacts on human emotions in the statistical analyses, we further explore a specific type of tourist attractions—\textit{amusement park} to illustrate the results. As shown in Table \ref{T:amusementparklist}, there are 17 amusement parks in this study and they generally have higher AHI (average 45.72) and Joy Index scores (average 0.52) compared with other types of tourist sites. For most amusement parks, they are located in urban  areas and with open space, as well as containing water bodies (e.g., lakes) inside the park. The average NDVI value at amusement parks is about 0.52, which is similar to the value of all sites (about 0.54) and not type-biased. A more specific exploration can be conducted in future to investigate more factors that may impact on the human emotions at amusement parks.

\begin{table}[H]
	\centering
	\caption{The coefficients of multi-linear regression based on the joy index and the geographical and environmental factors}
	\vspace{1em}
	\begin{tabular}{l l}
		\hline
		\textbf{Attributes} & \textbf{Regression Coefficient} \\
		\hline
		Constant & 0.486**\\
		Continent &  \\
		\textit{Asia}& -0.396*\\
		\textit{Europe}& -0.458**\\
		\textit{North America}& -0.353*\\
		\textit{Oceania}& -0.235 \\
		\textit{Africa}& N/A \\
		Open/Closed Space &  \\
		\textit{Open Space}& -0.0044 \\
		\textit{Closed Space}& N/A  \\
		Urban/Rural & \\
		\textit{Urban}& -0.1458* \\
		\textit{Rural}& N/A \\
		Type &  \\
		\textit{Cultural Landscape}& -0.1923** \\
		\textit{Museum}& -0.254*\\
		\textit{Natural Landscape}& 0.002 \\
		\textit{Palace}& -0.203* \\
		\textit{Religious Site}& -0.319* \\
		\textit{Amusement Park}& N/A  \\
		Water Body & \\
		\textit{Existence of the Water Body}& 0.021 \\
		\textit{Inexistence of the Water Body}& N/A  \\
		\textit{Distance to the Nearest Water Body}& 0.0004 \\
		NDVI & 0.0004 \\
		\hline
	\end{tabular}
	\\
	\(R^2 = 0.57$**$\) \\
	$*$ p $<$ 0.05 \\
	$**$ p $<$ 0.001 \\
	\label{T:multilinear_joyindex}
\end{table}

\begin{table}[H]
	\centering
	\caption{The coefficients of multi-linear regression based on the average happiness index (AHI) and the geographical and environmental factors.}
	\vspace{1em}
	\begin{tabular}{l l}
		\hline
		\textbf{Attributes} & \textbf{Regression Coefficient} \\
		\hline
		Constant & 60.649** \\
		Continent &  \\
		\textit{Asia}& -13.805* \\
		\textit{Europe}& -16.647* \\
		\textit{North America}& -12.196	 \\
		\textit{Oceania}& -5.058 \\
		\textit{Africa}& N/A \\
		Open/Closed Space &  \\
		\textit{Open Space}& -0.015 \\
		\textit{Closed Space}& N/A \\
		Urban/Rural & \\
		\textit{Urban}& -5.083* \\
		\textit{Rural}& N/A \\
		Type &  \\
		\textit{Cultural Landscape}& -9.264** \\
		\textit{Museum}& -10.645* \\
		\textit{Natural Landscape}& 0.81 \\
		\textit{Palace}& -10.889** \\
		\textit{Religious Site}& -15.547** \\
		\textit{Amusement Park}& N/A \\
		Water Body & \\
		\textit{Existence of the Water Body}& 0.7484 \\
		\textit{Inexistence of the Water Body}& N/A \\
		\textit{Distance to the Nearest Water Body}& 0.0172 \\
		NDVI & 0.018 \\
		\hline
	\end{tabular}
	\\
	\(R^2 = 0.57$**$\) \\
	$*$ p $<$ 0.05 \\
	$**$ p $<$ 0.001 \\
	\label{T:multilinear_ahi}
\end{table}

\begin{table}[H]
	\centering
		\caption{The list of amusement parks with their average happiness index (AHI) and joy index scores.}
	\vspace{1em}
	\begin{tabular}{|c|c|c|}
		\hline
		\textbf{Tourist site}              & \textbf{AHI} & \textbf{Joy Index} \\ \hline
		Epcot, USA                         & 53.86     & 0.60           \\ \hline
		Disney Animal Kingdom, USA         & 53.07    & 0.60         \\ \hline
		Disney Worlds Magic Kingdom, USA   & 52.91     & 0.60           \\ \hline
		Disney Holly Wood Studios, USA     & 50.37     & 0.56           \\ \hline
		Universal Studios, Hollywood, USA  & 48.05      & 0.54           \\ \hline
		Everland, Gyeonggi-Do, South Korea & 47.34     & 0.50          \\ \hline
		Disneyland Park, France            & 46.34     & 0.51          \\ \hline
		Disney California Adventure, USA   & 46.14     & 0.52          \\ \hline
		Ocean Park, Hong Kong              & 45.89     & 0.53           \\ \hline
		Disneyland Hong Kong, Hong Kong    & 45.76     & 0.51          \\ \hline
		Universal Studios, Florida, USA    & 45.55     & 0.53           \\ \hline
		Islands of Adventure, USA          & 45.34     & 0.52          \\ \hline
		Tokyo Disneyland, Japan            & 44.68     & 0.51         \\ \hline
		Universal Studios, Japan           & 41.57     & 0.47           \\ \hline
		Disneyland Park, USA               & 40.86    & 0.47          \\ \hline
		Balboa Park, USA                   & 38.6      & 0.45          \\ \hline
		Lotte World, South Korea           & 30.95     & 0.35           \\ \hline
	\end{tabular}
	\label{T:amusementparklist}
\end{table}
\section{Discussion}
\label{S:discussion}

\subsection{Human-environment perspective of results}
Scholars from environmental psychology have proved that surrounding environment has impacts on human emotions.
Results of this study demonstrate a similar conclusion from a big-data-driven perspective.
By combining the results of correlation analysis and the multiple linear regression, amusement parks are the places that most positively affect individual's happiness expressions.
Environments such as open spaces, places with the existence of water body, places where the green vegetables is denser and rural areas, seem to be summarized as one kind of places. 
All of these variables aforementioned present positive impacts on the degree of human happiness. 
Therefore, it can be concluded that people who stay in such areas may tend to feel happier. Our findings are consistent with several existing theory in psychology \citep{kaplan1995restorative}, that exposure to nature has a positive impact on human moods \citep{bowler2010systematic}, which also supports the theoretical foundation of the framework and proves the validity of this study to some extent.

However, some limitations should be pointed out. 
As expressions of human emotions are quite complex and are influenced by multiple variables both internally and externally, the results from this study may not be guaranteed for individuals \citep{junot2017passion} nor for all tourist attractions around the world.
And some cultural environment, religious sites and museums may suppress people's positive emotional expressions. It is worth noting that being suppressed does not mean that people are not happy at those places, but just express less enjoyment smiles explicitly.
In addition, although the semantics of geotags show that most photos are related to the places, tourists' emotions may not be directly relevant to the views of surrounding environments but could be affected by the activities they are doing or the events they are participating in at that place. A deeper exploration should be conducted to find out other factors impacting human emotion expressions.

\subsection{Uncertainty of the Data}

Social media data are uploaded by volunteers based on their experiences and opinions, which caused ``ambient geographic information" \citep{degrossi2018taxonomy}. 
As user-generated photos are used in this study, the uncertainty and quality of the data should be tested \citep{goodchild2012assuring}.
Three types of data uncertainty are addressed: the vagueness of the place, the different number of faces, and the different groups of people.

As the size and the boundary of a place might be vague, it is not suitable to use a fixed distance for data analysis.  Place boundaries are constructed based on the density distribution of photos. Georeferenced photos outside the place boundary are removed to minimize the error of the results.
In addition, by using the DBSCAN algorithm, which is not sensitive to the noise data for place construction, the vagueness of the results are also decreased.
Besides, a combination of parameter settings as well as the Kendall’s W are tested for ensuring data consistency. Therefore, the uncertainty of the result is minimized.

Since the number of faces varied across different tourist sites, a key issue is to examine whether the number of photos collected at one site is sufficient to extract the human emotion and whether the emotional condition calculated is stable. Using the bootstrapping strategy, a 95\% confidence interval of emotion scores at each site is generated. The variability is derived by subtracting the lower bound from the upper bound of the confidence interval. To explore the relationship between the uncertainty of the emotional indices and the number of faces analyzed at each site, the linear regression analysis was employed. Figure \ref{F:bootstrap_joy_happiness} illustrates that the relation between the variability of emotional indices at each site and the number of faces identified from photos taken at each site fits into a power model very well (with a goodness of fit coefficient 0.99). In general, the more faces detected at a site, the more stable is the emotion measurement calculated. For most sites, the variability of 95\% confidence interval for the joy index is less than 0.05 and is less than 3 for the AHI, and have limited influence on the ranking lists. Therefore, the results of the emotional conditions at each site are reliable and can be trusted.

In addition, as different groups of people with various culture backgrounds and being locals or visitors may express different degree of excitement, enjoyment, and emotions to the same place, the results might be influenced by the proportion of various types of tourists. For instance, in order to distinguish the tourists and local people, we follow the criteria used in previous studies to define tourists:
 if the period of one user who takes multiple photos at one place longer than one month, then the user was labeled as locals otherwise as visitors \citep{garcia2015identification}.
Results show that for most tourist attractions (more than 90\%), the majority of Flickr photos (more than 80\%) are uploaded by tourists.
The average difference of the AHI scores between tourists and locals in those tourist sites is just 3, showing that such influence is minimal and will not change the ranking list.

Although we tried our best to reduce the uncertainty, some limitations still exist. Data bias commonly exists in the VGI data \citep{senaratne2017review}. 
As suggested previously \citep{gao2017constructing,jolivet2017crowd}, one data bias issue of VGI is that the contributions of volunteers often follow a power-law or an exponential-law of frequency distribution with a long tail, which indicates that most photos are posted by only a small proportion of users and a large number of users only contribute few \citep{goodchild2012assuring}. 
In this study, a large number of faces detected might belong to a small group of users. And the information provided by social media users may not always comply with quality standards. However, the results of emotions based on facial expressions do reflect active users' experiences, opinions, interests, and feelings at those places, and can provide new insights for the place-based information research \citep{blaschke2018place}.

\afterpage{
	\begin{figure}[H]
		\centering\includegraphics[width=0.73\linewidth]{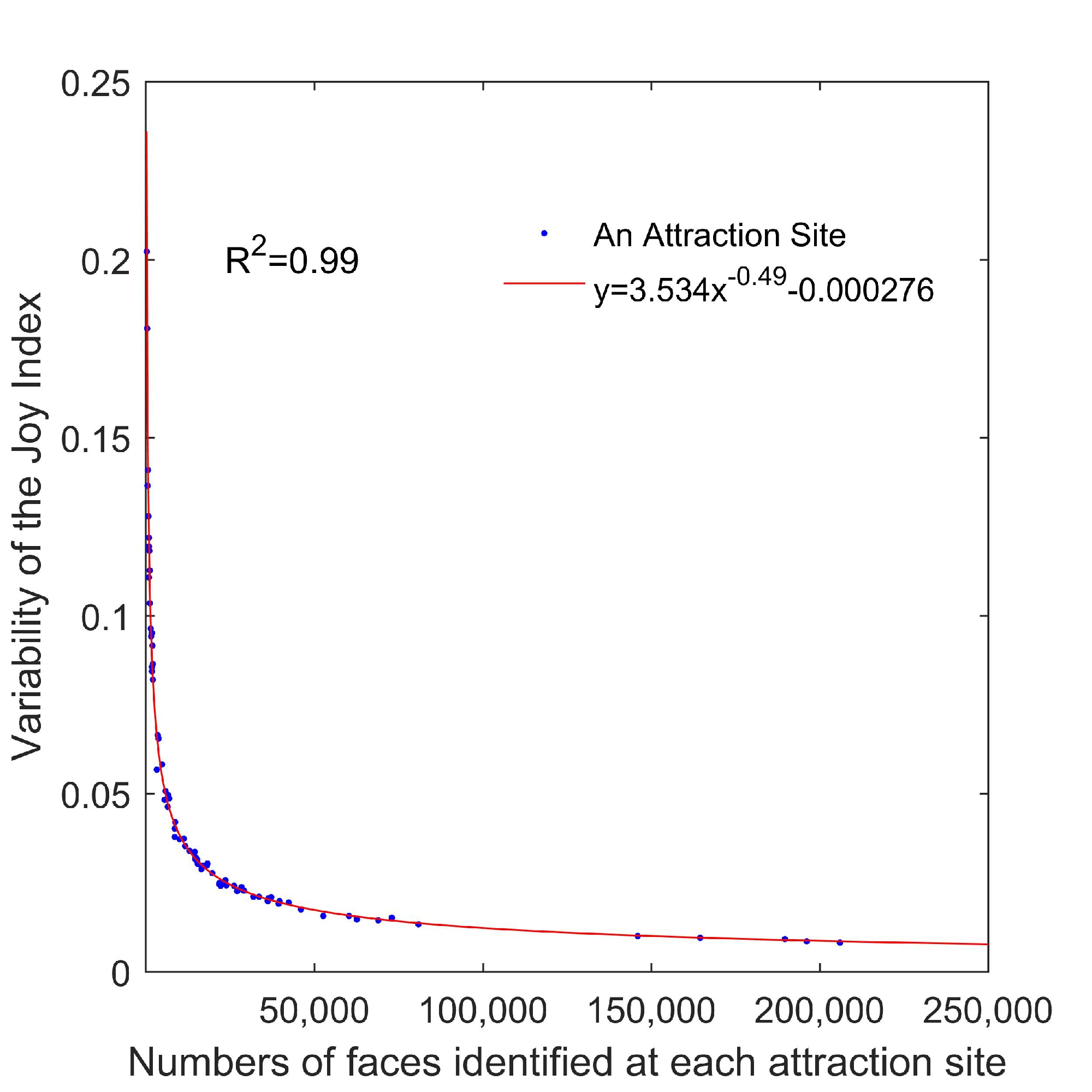}
		\\(a)\\
		\centering\includegraphics[width=0.73\linewidth]{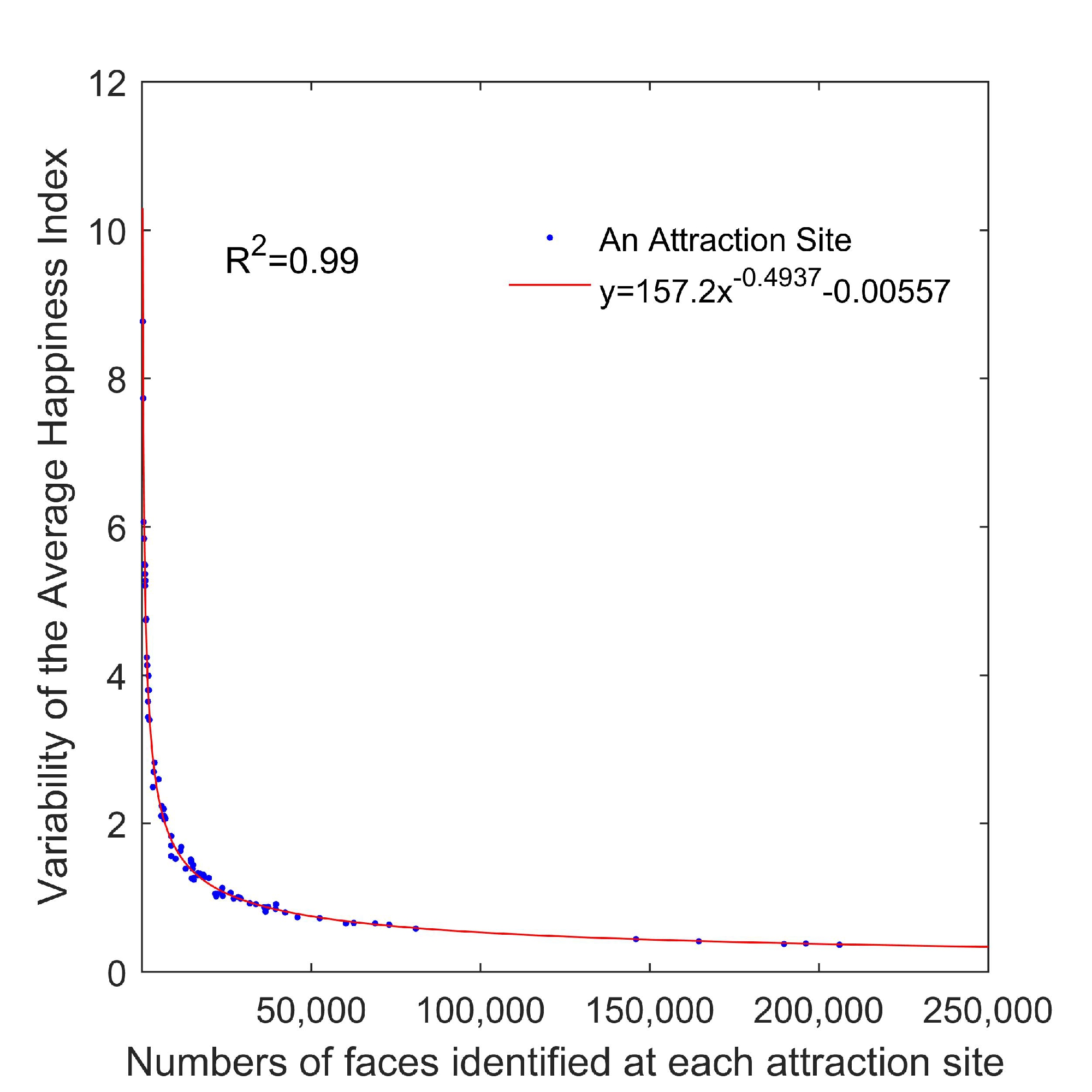}
		\\(b)\\
		\caption{The relation between the variability of emotional indices at each site and the number of faces identified from photos based on: (a) Joy Index, (b) Average Happiness Index.}
		\label{F:bootstrap_joy_happiness}
	\end{figure}
	\clearpage}

\subsection{Comparison between text-based and facial-expression-based methods}
Though facial-expression based emotional detection methods have becoming more mature, and a few of studies implemented it in research, a key issue is whether the facial-expression based methodology is reliable.
Therefore, we conducted a comparison between our methodology and a text-based framework.
To address this, we referred to the Mitchell's research \citep{mitchell2013geography}.
In this study, scholars followed Dodd's method \citep{dodds2011temporal}, where a daily happiness score was calculated from Twitter with the state-of-the-art NLP technologies, summarizing a range of human emotions in United States from a state level, and examining the connections to the socioeconomic attributes.
In comparison, the YFCC100 dataset containing the most photos in Flickr was used \citep{thomee2015yfcc100m}.
We also evaluated emotions using our framework of all photos in each state in United States.
Since Mitchell's research was conducted in the year 2011, we only retrieved photos taken in the same year within the United States to ensure the consistence of the time period.
Then, both Joy Index and Average Happiness Index were applied for the photo data to calculate the happiness scores for each state.
The results of our metrics and the results of the Mitchell's research across 50 states were analyzed via the Spearman's correlation analysis \citep{fieller1957tests}, which makes comparison between the rank of each value in the data series.
As shown in Table \ref{T:text-image}, results in the two studies have positive correlation: the Joy Index 0.28 and the Average Happiness Index 0.30, which show some degree of similarity between the two technologies.

It should be noted that the focus of our research is not to compare and even contrast the existing text-based emotion extraction technologies with facial expression approaches.
Different research methods have their own pros and cons. 
As aforementioned, text-based approaches cannot record real-time emotions and might not be suitable for global-scale research due to the multi-linguistic environment.
But it typically has larger volumes of datasets and rich semantics \citep{hu2019placesentiment}.
By combining those two approaches for affective computing could help improve the holistic understanding of human emotions from different aspects and enrich the understanding of such innate neural program \citep{abdullah2015collective}.

Our approach also has some limitations. First, as aforementioned, the results might be biased to certain group of people (visitors v.s. local citizens and different ethnicities or culture backgrounds) and affected by the diversity of faces in the trained data sets. Second, people's emotions may not be directly relevant to the views of surrounding environments. Moreover, people may not always express emotions explicitly by either facial expressions nor texts.
A further exploration should be conducted in order to show the collective connections between human emotions and the facial expressions in the technology-mediated platforms \citep{elwood2015technology}.


\begin{table}[H]
\centering
\caption{The Spearman's correlation coefficients between the text-based method and the facial expression-based method.}
\vspace{1em}
\begin{tabular}{l l l}
\hline
\textbf{Emotion Index} & \textbf{Correlation Coefficient} & \textbf{p-value}\\
\hline
Joy Index & 0.28 & 0.0472\\
Average Happiness Index & 0.30 & 0.0314\\
\hline
\end{tabular}
\label{T:text-image}
\end{table}

\section{Conclusion and Future Work}
\label{S:conclusion}

In this research, in order to understand the interaction between human emotions and the environment, a data-driven framework of measuring the human emotions at different places using large-scale user generated photos from social media is proposed.
We utilize the state-of-the-art social computing tools to detect and measure human happiness from facial expressions in photos. Tourist attractions, as a specific type of place are exemplified for deriving place-based human emotion indices.
A ranking list of 80 tourist sites across the world is created not from the statistics of tourist flow, but from the degree of happiness expressed and shared by millions of tourists, which also shows that our framework is suitable for global-scale issues.
In addition, we explore the impacts of several geographical and environmental factors to human happiness. Results are consistent with common sense and with existing studies from psychology, stating that people in the environment with more openness and with more opportunities exposing to nature express more happiness and smiling on faces.
Overall, this research advances our knowledge of the human emotions at different tourist attractions. Our study connects crowdsourced human emotions to the geographic attributes of environment using advanced AI techniques and spatial analytics, and provides a new paradigm for research in geography and in GIScience. The proposed framework and the findings could also lead to practical guidance for environmental psychology, human geography, tourism management and urban planning. 

In the future, several potential directions will be focused to explain related research questions. One direction is about data fusion. Although only Flickr photos are employed in the experiment, this study can be further improved with diverse data sources such as surveys. A data-synthesis-driven method might provide varied perspectives of human emotions. And the mix of text-based and facial expression-based emotion extraction methods may enhance the confidence of the final output. Another direction is to explore fundamental factors impacting human emotions. As we propose the framework for \textit{Place Emotion} research, we will focus more on spatial analysis of the emotion patterns. Human emotions at different scales will be compared to revisit the scale effect in geography. And different groups of people, as suggested by existing studies \citep{niedenthal2018heterogeneity, kang2018happier}, will be explored to figure out deeper insights on influential factors of human emotions. 
Moreover, different place types as well as spatial units from different scales, including points of interest, census blocks, neighborhoods, and communities will be combined to examine the geographic patterns and socioeconomic linkages of human emotions.
One specific research taking a limited number of places but with more environmental and socioeconomic factors to be examined can be conducted to enrich the understanding of place-based emotions.

\section*{Acknowledgement}
The authors would like to thank Wanjuan Bie, Shan Lu, and Dan'nan Shen at Wuhan University, for their contributions on figures. Thank Timothy Prestby at UW-Madison for his help on language edits of the manuscript. 
And thank Jialin Wang, Zimo Zhang, Wenyuan Kong and Zijun Xu in the Place\&Emotion Group, Urban Playground Lab, Wuhan University, for their helpful discussions.
The funding support for this research is provided by the Office of Vice Chancellor for Research and Graduate Education at the University of Wisconsin-Madison with funding from the Wisconsin Alumni Research Foundation, and the Fund for National College Students Innovations Special Project of China (Grant No. 201810486033).

\bibliographystyle{model1-num-names}
\bibliography{ref}

\end{document}